\documentclass{article}

     \usepackage[final,nonatbib]{neurips_2021}

\usepackage[utf8]{inputenc} 
\usepackage[T1]{fontenc}    
\usepackage{url}            
\usepackage{booktabs}       
\usepackage{amsfonts}       
\usepackage{nicefrac}       
\usepackage{microtype}      
\usepackage{xcolor}         
\usepackage{graphicx}
\usepackage{caption}
\usepackage{subcaption}

\definecolor{mydarkblue}{rgb}{0,0.08,0.45}
\usepackage[colorlinks=true,
    linkcolor=mydarkblue,
    citecolor=mydarkblue,
    filecolor=mydarkblue,
    urlcolor=mydarkblue]{hyperref}  

\usepackage{hyperref}    
\makeatletter
\newcommand{\printfnsymbol}[1]{%
  \textsuperscript{\@fnsymbol{#1}}%
}

\graphicspath{{Figures/}}

\title{Generating Diverse Realistic Laughter\\ for Interactive Art}
\author{
  M. Mehdi Afsar\thanks{Equal contribution, corresponding author: \texttt{mehdi.afsar@ucalgary.ca}}\\
 Mila \& University of Calgary\\
  \And
  Eric Park\footnotemark[1]\\
  Mila \& University of Waterloo\\
\And
Étienne Paquette\\
Independent Artist\\
  \And
  Gauthier Gidel\thanks{Canada CIFAR AI Chair}\\
  Mila \& University of Montreal\\
  \And
  Kory W. Mathewson\\
  DeepMind\\
  \And
  Eilif Muller\footnotemark[2]\\
  Mila \& University of Montreal\\
}

\begin{document}

\maketitle

\begin{abstract}

We propose an interactive art project to make those rendered invisible by the COVID-19 crisis and its concomitant solitude reappear through the welcome melody of laughter and connections created and explored through advanced laughter synthesis approaches.
However, the unconditional generation of the diversity of human emotional responses in high-quality auditory synthesis remains an open problem with important implications for applying these approaches in artistic settings. We developed LaughGANter, an approach to reproduce the diversity of human laughter using generative adversarial networks (GANs).  When trained on a dataset of diverse laughter samples, LaughGANter generates diverse, high-quality laughter samples and learns a latent space suitable for emotional analysis and novel artistic applications such as latent mixing/interpolation and emotional transfer.\footnote{Samples: \href{https://bit.ly/2XQqcW0}{https://bit.ly/2XQqcW0}}
\end{abstract}
\vspace{-9pt}
\section{Introduction}


\vspace{-8pt}
Modern society has refined the condition of solitude to the point where countless seniors, marginalized because of their age, have magically disappeared: left to their own devices, these individuals fade from social life and essentially live in a parallel world. 
The COVID-19 crisis and resulting lockdowns have both entrenched this phenomenon and helped reveal its widespread. Can artificial intelligence (AI) help reconnect generations by making them part of a transgenerational art experience? At the crossroads of laughter—an act of communication between two individuals—and artificial intelligence—a purely functional entity—can we rediscover our humanity?
\vspace{-8pt}

\paragraph{An interactive experience.}
The end goal of this project is to connect people via an interactive web experience driven by synthetic laughter. Using our models, we
will explore the phenomenon of empathy triggered by laughter, the relationship between individual memory and laughter, and how the sound of laughter evolves over a lifetime.
\vspace{-8pt}
\paragraph{Laughter generation for advancing audio synthesis research.}
 With stunning advancements in image synthesis~\cite{karras2017progressive, karras2019style, karras2020analyzing, karnewar2020msg}, Generative Adversarial Networks (GANs) ~\cite{goodfellow2014generative} have gained the attention of researchers in the field of audio synthesis~\cite{donahue2018adversarial, engel2019gansynth, kumar2019melgan, binkowski2019high}.  Synthesizing audio opens new doors for musicians and artists and enables them to expand their repertoire of expression~\cite{donahue2018adversarial}. 
 Despite significant progress by the ML community on methods for audio synthesis, there have been only a few attempts in the topic of laughter synthesis~\cite{mancini2013laugh}, and none leveraging modern approaches such as GANs.
 

Compared to speech, laughter is made challenging by its many context-dependent attributes, such as emotions~\cite{schroder2001emotional}, age, and gender.  Moreover, compared to well-studied topics like speech synthesis, there are no established evaluation methods for synthesized laughter.
Thus laughter synthesis has the potential to become a standard benchmark in unconditional audio synthesis. 

\vspace{-8pt}
\paragraph{Related work.}
 Previous work in the field of laughter generation involves~\cite{mori2019conversational} the use of oscillatory system~\cite{sundaram2007automatic}, formant synthesis~\cite{oh2013lolol}, articulatory speech synthesis~\cite{lasarcyk2007imitating}, and hidden Markov models (HMM)~\cite{urbain2014arousal}.  Recently, some researchers have also used deep learning~\cite{mori2019conversational, tits2020laughter} methods for laughter synthesis. However, GANs are advantageous in learning a compact latent space allowing for interpolation, mixing, and style transfer as well as emotional analysis. In this paper, we propose to use GANs for unconditional laughter generation and manipulation (LaughGANter). We aim to enable a unique interactive art experience that surprises and connects  through the primordial intimacy of our laughter interacting and juxtaposing with others.

\vspace{-8pt}
\section{Methodology}
\vspace{-8pt}
We adapt Multi-Scale Gradient GAN (MSG-GAN)~\cite{karnewar2020msg} for laughter synthesis. Among other popular image synthesis methods, like DCGAN~\cite{radford2015unsupervised}, ProgressiveGAN~\cite{karras2017progressive}, and StyleGAN~\cite{karras2019style}, LaughGANter employs multi-scale gradients
on a DCGAN architecture to address the training instability prevalent in GANs. As a result, progressive growing of network resolutions is avoided to limit the hyperparameters to be tuned (e.g., training schedule, learning rates, etc.). At the same time, the multi-scale discriminator penalizes the generator's intermediate and final layer outputs.


We refer the reader to~\cite{karnewar2020msg} for an in-depth study of the MSG-GAN architecture.  Concisely, the generator ($G$) samples a random vector $z$ from a normal distribution and outputs $x=G(z)$. The generated samples are fed into the discriminator ($D$), along with real samples, to measure the divergence. We perform \textit{pixel normalization} after every layer in $G$, and employ the \textit{Relativistic Average Hinge} loss~\cite{jolicoeur2018relativistic} in $D$. Moreover, inspired by~\cite{oord2016wavenet, odena2016deconvolution}, we explored the impact of \textit{induced receptive field expansion}, adding residual blocks with dilations after each upsampling layer in $G$, exponentially increasing the model's receptive field and leading to better long-range correlation in audio data. 
\vspace{-8pt}
\paragraph{Categorical Conditional Generation.}
A more directed data generation process is employed through a conditional adaptation of MSG-GAN ~\cite{mirza2014conditional}, facilitating the laughter representation learning given additional context beyond unlabeled laughter (e.g., gender, age, humor style, etc.). Here, categorical information augments the latent noise vector in $G$, and to each multi-scale vector within $D$, through concatenation with an embedding of context information. 

\vspace{-10pt}

\begin{figure}[t]
	\label{fig:spec}
	\centering
	\begin{subfigure}[b]{0.32\textwidth}
	\centering
    	\includegraphics[width=\textwidth]{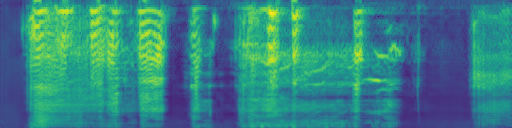}
    	\vspace{.5cm}
    	\caption{\small Generated Mel spectrogram}
	\end{subfigure}
	\hspace{2pt} 
	\begin{subfigure}[b]{0.32\textwidth}
    	\includegraphics[width=\textwidth]{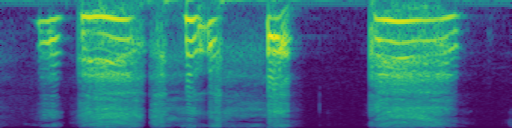}
    	\vspace{.5cm}
    	\caption{ \small Real Mel spectrogram}
    \end{subfigure}
    \hspace{2pt}
	\begin{subfigure}[b]{0.32\textwidth}
	\centering
    	\includegraphics[width=\textwidth]{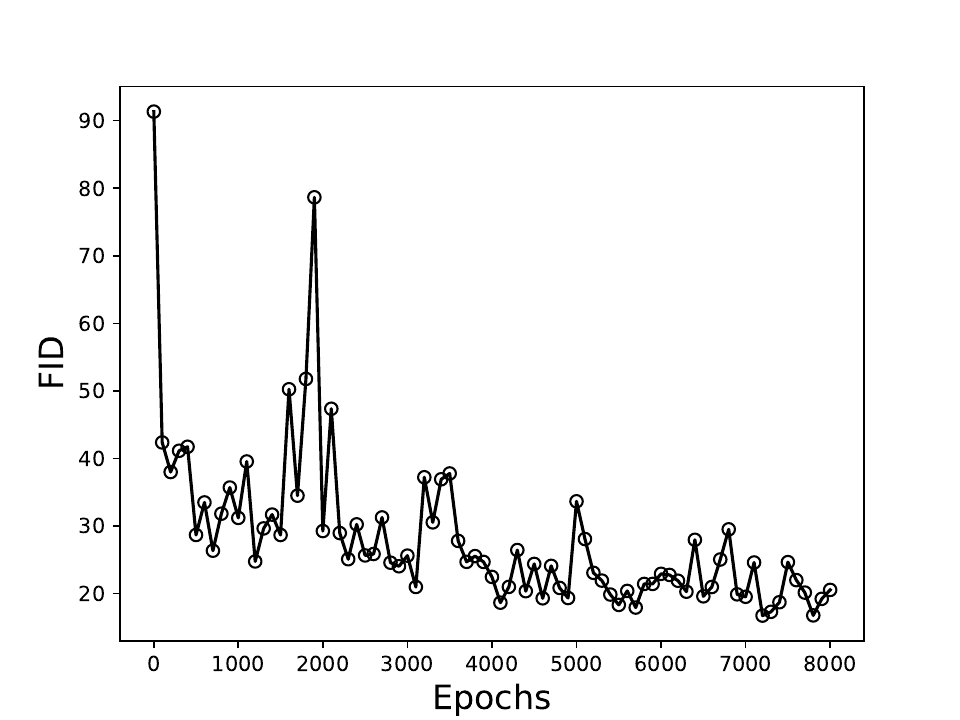}
    	\caption{\small  FID score for LaughGANter}
	\end{subfigure}
 	\caption{(a)-(b) Log-magnitude spectrograms for real and generated samples show similar features on qualitative analysis, and (c) FID score for LaughGANter decreases during training as the diversity of the generated samples approaches that of the training data distribution.}
	\label{fig:results}
\end{figure}

\section{Experiments}
\vspace{-8pt}
\paragraph{Setup.}
Our model is implemented in PyTorch. We use a dataset containing 2145 laughter samples collected by the National Film Board of Canada. Samples are 1-8s long (22.05kHz  mono) and were organized (and labeled) from subjects of different ages and genders (55\% male, 45\% female; 93\% adult, 6\% child, 1\% teen). The audio data is augmented using  a random combination of additive noise, shifting, and changing pitch and duration (using \texttt{pyrubberband}). Then, this data is converted to Mel spectrograms and fed into the model. In addition to qualitative evaluation, i.e., listening to generated samples, we have used Fréchet inception distance (FID)~\cite{heusel2017gans} to assess the diversity of the generated samples compared to the training dataset. Instead of using Inception features used in the original FID score, we use features from a classifier (gender and age group) trained on the spectrograms of our laughter dataset.
\vspace{-8pt}

\clearpage



\bibliography{general}
\bibliographystyle{unsrt}

\clearpage
\section*{Supplementary Materials}

\subsection*{Ethical Implications}
LaughGANter's facilitation in the rediscovery of our humanity and re-connectivity of generations through interactive art experiences allows for the production of a diversity of human emotional responses. Such a system's capability in emotional transfer can be used to strengthen the interrelationship within human-and-machine interaction, however, it is capable of coaxing accurate and precise emotional responses, which could be used for downstream human manipulation tasks. 
This project is explicitly performed for artistic purposes, so ethical considerations applied to all creative projects also apply to our work.

\subsection*{Acknowledgements}

We want to thank Arnaud Roussel for his contributions to dataset processing and early MSG-GAN prototypes, as well as Isabelle Repelin, Isabelle Limoges, Martin Viau, Stephanie Quevillon, and Marie-Eve Babineau at the National Film Board of Canada for financial and project management support for this research.

\subsection*{Experiments Results}
Fig.~2 depicts how the performance of LaughGANter is improved in the course of training. In particular, Figs.~2(k) and (l) show that LaughGANter can generate laughter samples that are very close to real samples.  Moreover, Fig.~3 depicts a linear interpolation between the generated laughter of a female and a male.

\begin{figure}[h!]

	\centering
	\begin{subfigure}[b]{0.3\textwidth}
    	\includegraphics[width=\textwidth]{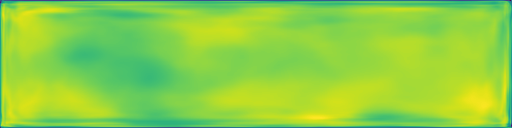}
    	\caption{Epoch 1}
	\end{subfigure}
	\begin{subfigure}[b]{0.3\textwidth}
    	\includegraphics[width=\textwidth]{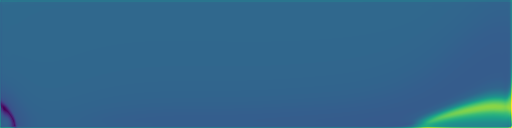}
    	\caption{Epoch 10}
	\end{subfigure}
		\begin{subfigure}[b]{0.3\textwidth}
    	\includegraphics[width=\textwidth]{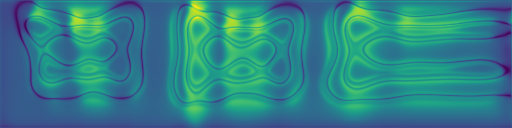}
    	\caption{Epoch 100}
	\end{subfigure}
	\begin{subfigure}[b]{0.3\textwidth}
    	\includegraphics[width=\textwidth]{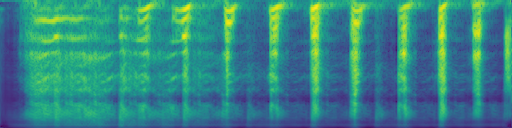}
    	\caption{Epoch 1000}
    \end{subfigure}
	\begin{subfigure}[b]{0.3\textwidth}
    	\includegraphics[width=\textwidth]{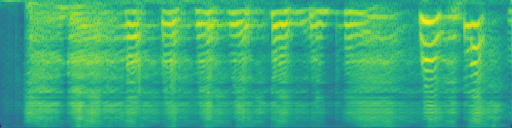}
    	\caption{Epoch 2000}
	\end{subfigure}
		\begin{subfigure}[b]{0.3\textwidth}
    	\includegraphics[width=\textwidth]{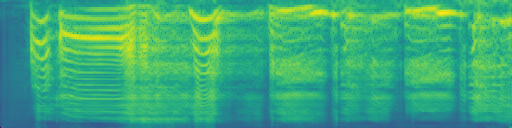}
    	\caption{Epoch 3000}
    \end{subfigure}
    \begin{subfigure}[b]{0.3\textwidth}
    	\includegraphics[width=\textwidth]{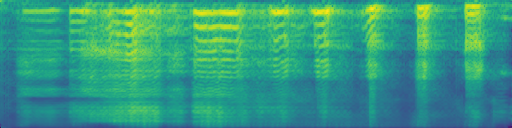}
    	\caption{Epoch 4000}
    \end{subfigure}
    \begin{subfigure}[b]{0.3\textwidth}
    	\includegraphics[width=\textwidth]{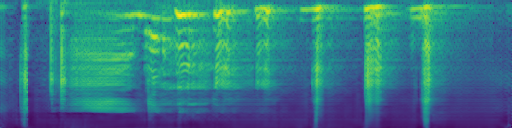}
    	\caption{Epoch 5000}
    \end{subfigure}
        \begin{subfigure}[b]{0.3\textwidth}
    	\includegraphics[width=\textwidth]{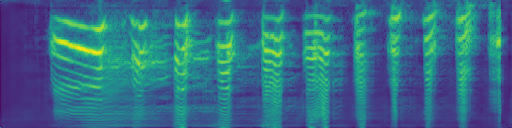}
    	\caption{Epoch 6000}
    \end{subfigure}
        \begin{subfigure}[b]{0.3\textwidth}
    	\includegraphics[width=\textwidth]{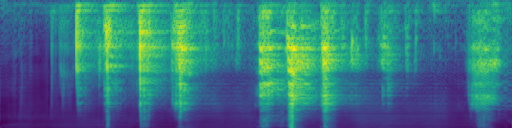}
    	\caption{Epoch 7000}
    \end{subfigure}
            \begin{subfigure}[b]{0.3\textwidth}
    	\includegraphics[width=\textwidth]{spec8000.png}
    	\caption{Epoch 8000}
    \end{subfigure}
    \begin{subfigure}[b]{0.3\textwidth}
    	\includegraphics[width=\textwidth]{real1.png}
    	\caption{A real sample}
    \end{subfigure}
\label{fig:spec}
\caption{(a)-(k) Log-magnitude spectrograms of generated laughter samples during training and (l) a real sample }
\end{figure}

\begin{figure}[t]
	\label{fig:spec}
	\centering
	\begin{subfigure}[b]{0.3\textwidth}
    	\includegraphics[width=\textwidth]{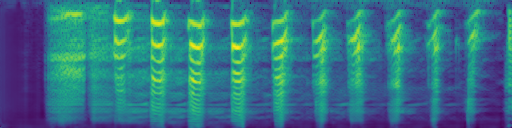}
    	\caption{}
	\end{subfigure}
	\begin{subfigure}[b]{0.3\textwidth}
    	\includegraphics[width=\textwidth]{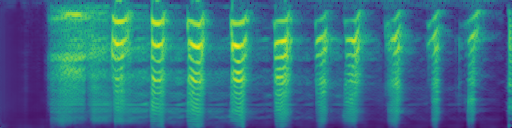}
    	\caption{}
	\end{subfigure}
		\begin{subfigure}[b]{0.3\textwidth}
    	\includegraphics[width=\textwidth]{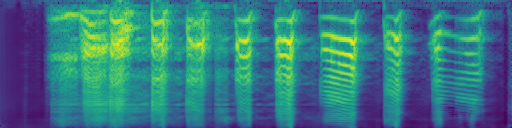}
    	\caption{}
	\end{subfigure}
	\begin{subfigure}[b]{0.3\textwidth}
    	\includegraphics[width=\textwidth]{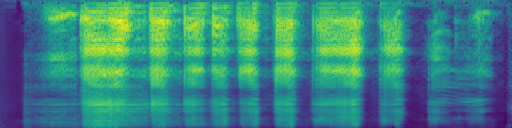}
    	\caption{}
    \end{subfigure}
	\begin{subfigure}[b]{0.3\textwidth}
    	\includegraphics[width=\textwidth]{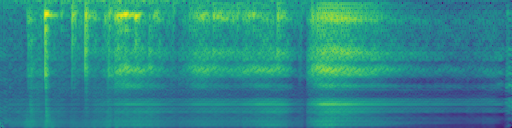}
    	\caption{}
	\end{subfigure}
		\begin{subfigure}[b]{0.3\textwidth}
    	\includegraphics[width=\textwidth]{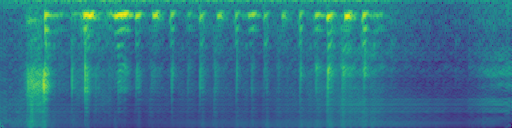}
    	\caption{}
    \end{subfigure}
    \begin{subfigure}[b]{0.3\textwidth}
    	\includegraphics[width=\textwidth]{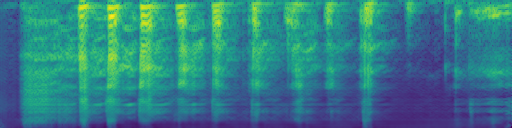}
    	\caption{}
    \end{subfigure}
    \begin{subfigure}[b]{0.3\textwidth}
    	\includegraphics[width=\textwidth]{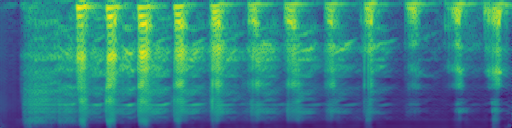}
    	\caption{}
    \end{subfigure}
        \begin{subfigure}[b]{0.3\textwidth}
    	\includegraphics[width=\textwidth]{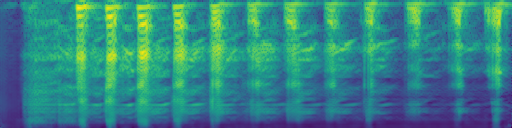}
    	\caption{}
    \end{subfigure}
        \begin{subfigure}[b]{0.3\textwidth}
    	\includegraphics[width=\textwidth]{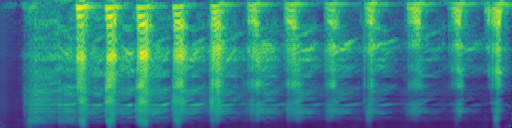}
    	\caption{}
    \end{subfigure}
\caption{Interpolation between generated laughter respectively identified as female (a) and  male (j)}
\end{figure}

\end{document}